\documentclass[a4paper]{article}

\usepackage{INTERSPEECH2022}
\usepackage{svg}
\usepackage[para]{manyfoot}
\usepackage[export]{adjustbox}
\usepackage{url}

\usepackage{lipsum}

\newcommand\blfootnote[1]{%
  \begingroup
  \renewcommand\thefootnote{}\footnote{#1}%
  \addtocounter{footnote}{-1}%
  \endgroup
}

\title{ByT5 model for massively multilingual grapheme-to-phoneme conversion}
\name{Jian Zhu$^1{^*}$, Cong Zhang$^2{^*}$, David Jurgens$^3$}

\address{
  $^1$Department of Linguistics, University of Michigan, Ann Arbor, USA\\
  $^2$Center for Language Studies, Radboud University, Nijmegen, Netherlands \\
  $^3$School of Information, University of Michigan, Ann Arbor, USA }
\email{lingjzhu@umich.edu, cong.zhang@ru.nl,jurgens@umich.edu}

\begin{document}
\maketitle

\begin{abstract}
In this study, we tackle massively multilingual grapheme-to-phoneme conversion through implementing G2P models based on ByT5. We have curated a G2P dataset from various sources that covers around 100 languages and trained large-scale multilingual G2P models based on ByT5. We found that ByT5 operating on byte-level inputs significantly outperformed the token-based mT5 model in terms of multilingual G2P. Pairwise comparison with monolingual models in these languages suggests that multilingual ByT5 models generally lower the phone error rate by jointly learning from a variety of languages. The pretrained model can further benefit low resource G2P through zero-shot prediction on unseen languages or provides pretrained weights for finetuning, which helps the model converge to a lower phone error rate than randomly initialized weights. To facilitate future research on multilingual G2P, we make available our code and pretrained multilingual G2P models at: \url{https://github.com/lingjzhu/CharsiuG2P}.
\end{abstract}

\section{Introduction}
Grapheme-to-phoneme conversion, commonly known as G2P, is the task of converting orthographic symbols (\textit{graphemes}) in a language into phonetic symbols (\textit{phonemes}). 
G2P is a fundamental to the pipeline for a variety of speech processing tasks that depend on phonemic inputs, including speech synthesis and speech recognition. While G2P has been a well researched area for a few high resource languages \cite{bisani2008joint,novak-etal-2012-wfst,rao2015grapheme,yao15_interspeech,toshniwal2016jointly,mousa2016deep,milde2017multitask,yolchuyeva19_interspeech,Sun2019,Park2020}, G2P tools are still lacking for most languages.
\blfootnote{* Equal contributions.}
With the advent of massive multilingual speech models like XLS-R \cite{babu2021xls,conneau2020unsupervised}, multilingual pretraining has become a potential method to allow for resource speech processing. While multilingual speech recordings are becoming ever more available, the speech processing pipeline often relies on phonemic transcriptions. Therefore, the availability of multilingual G2P toolkits will greatly facilitate multilingual speech processing.

Multilingual G2P is an active area of research \cite{deri-knight-2016-grapheme,peters-etal-2017-massively,hermjakob-etal-2018-box,hasegawa2020grapheme,gorman-etal-2020-sigmorphon,yu2020multilingual,vesik-etal-2020-one}. Multilingual models increase efficiency by training a single model that can process multiple languages, rather than training a separate model for each language. While this multilingual training process is more challenging, multilingual pretrained transformers have been closing the gap between multilingual models and monolingual models. 
Despite the potential benefits, there are relatively few publicly-available tools for multilingual G2P, notably \texttt{Epitran} \cite{mortensen-etal-2018-epitran}, \texttt{Phonetisaurus} \cite{novak2016phonetisaurus}, and \texttt{eSpeak-NG} \cite{espeakng}. These rule-based or finite-state transducer (FST) based models work well for many languages but they still leave substantial room for improvement in both covering more languages and improving the G2P accuracy.

Yet multilingual G2P still faces many non-trivial barriers, both in terms of data and of models. One is the lack of multilingual pronunciation dictionaries to train multilingual models. Moreover, world languages have a wide range of writing systems, how to encode a huge number of orthographic symbols in neural models remains challenging. To tackle multilingual G2P problem, we address both challenges. First, to create a training dataset, we aggregated pronunciation dictionaries previously published or made available in around 100 languages. Second, to encode diverse writing systems, we trained on 99 languages the sequence-to-sequence ByT5 model that takes raw bytes as processing units. Our results show that byte-level ByT5 outperformed the token-based mT5 models in multilingual G2P with far fewer parameters and the multilingual ByT5 also outperformed most \textit{mono}lingual G2P models with the same architecture. Moreover, multilingual models can perform zero-shot G2P on unseen low-resource languages with seen writing systems. Pretrained weights can also be fine-tuned on low resource languages to speed up convergence and increase performance. Our proposed method represent an efficient strategy for multilingual and low-resource G2P problems and we make our models publicly available to facilitate future research.

\section{Multilingual Pronunciation Dictionaries}

\begin{figure*}[tbh]
    \centering
    \includegraphics[width=\textwidth]{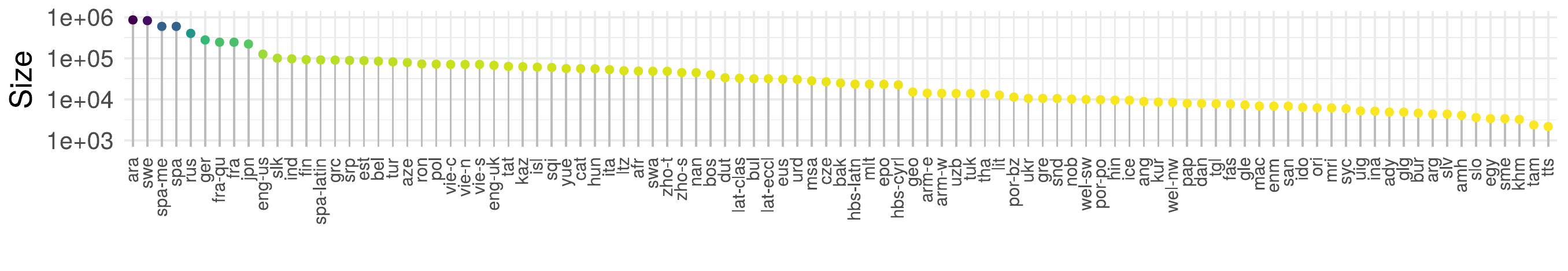}
    \caption{Distribution of pronunciation dictionary sizes for 99 languages in the training set. Languages are shown as ISO-639 codes.}
    \label{fig:distribution}
\end{figure*}

In order to create a large pronunciation dataset, we manually selected and merged several publicly available G2P datasets \cite{gorman-etal-2020-sigmorphon,lee-etal-2020-massively} and constructed G2P mapping dictionaries for some new languages. To promote future research, we released a detailed catalogue of available G2P data and their licenses at: \url{https://github.com/lingjzhu/CharsiuG2P\#g2p-datasets}.

\textbf{Collecting existing dictionaries}
Pronunciation dictionaries were collected from various sources, including online dictionaries for individual language varieties \cite{gorman-etal-2020-sigmorphon, veisi2020toward,  britfone, Sprakbanken_Swe, sandiago-spanish}, and dictionary collections such as \textit{ipa-dict} \cite{ipa-dict} and \textit{Wikipron} \cite{lee-etal-2020-massively}. Dictionaries from the same variety was merged. 
Repeated entries were removed and different pronunciations for the same word were kept. We made our best effort to select only data with licenses that do not restrict their usage.

Pronunciation dictionaries for individual languages were prioritized over Wikipron for their higher accuracy. If Wikipron was the only source for a language, the dictionaries in Wikipron (e.g. dictionaries with narrow vs broad transcription) were then compared. When a Wikipron dictionary had a substantially larger number of entries than another, this dictionary was selected. When two Wikipron dictionaries had similar numbers of entries, the one with narrow transcriptions was selected.

\textbf{Creating pronunciation dictionaries}
For languages without an existing pronunciation dictionaries or only have a small number of entries in Wikipron, we created the pronunciation dictionaries with \texttt{espeak-NG}. For each language, we obtained a wordlist of that language from the Leipzig corpora \cite{goldhahn2012building}, which hosted text data in more than 200 languages. A frequency threshold of 10 was set to filtered out low frequency words and noisy data. Words with numbers, punctuation or non-character symbols were all removed but words with IPA diacritics were all kept. These clean wordlists were converted into pronunciation dictionaries through \texttt{espeak-NG}.

\subsection{Data partitioning}
Languages were coded following the ISO 639-3 convention. When a language variety did not have an available ISO 639-3 code, a triple letter ISO 639-2 or 639-5 code was used \cite{iso_download}. If more than one variety of the same language were included, a suffix representing the variety was added, e.g. \texttt{eng-uk} and \texttt{eng-us}. Figure~\ref{fig:distribution} illustrates the distribution of dictionary sizes for each language. 

After preprocessing, we ended up with 99 languages with more than 3000 entries. 
We constructed the development and the test sets by setting aside 50 words and 500 words from each language dictionary respectively. The rest of the data were used as training data, which totaled 7.2 million words across all languages in the data. In addition, we also selected extra 5 low-resource languages to test G2P in low resource settings, which will be discussed in detail at Section~\ref{low}.

As pronunciation dictionaries only contain mappings between words and their IPA transcription, the current setting is only suitable for training word-level G2P models. While homographs in many languages require phrasal contexts to correctly pronounce (e.g. present vs. past tenses of `read' in English), this is not a primary focus of this study. Phonemic transcriptions for sentences are only available for a very small subset of high resource languages like English and Chinese. For the majority of languages, even word-level G2P tools are lacking, so we consider developing G2P tools for these languages are still necessary.

\section{Modeling G2P}
G2P is a sequence transduction task in which an input orthographic sequence $\boldsymbol{W}=[w_1,w_1,\dots,w_n]$ is transformed to an output phoneme sequence $\boldsymbol{P}=[p_1,p_2,\dots,p_t]$ through a mapping function $f$, which is a neural network in this study. To tackle challenges in multilingual G2P, we implemented neural models with the following considerations. 

\textbf{Byte-level encoding}  For language modeling, the common approach to encode texts is to employ the byte pair encoding (BPE) to segment raw texts into subword units \cite{sennrich-etal-2016-neural}, which are then mapped to a set of embeddings. However, for multilingual models, BPE tokenizaation could potentially result in a huge vocabulary size, inflating model parameters. In contrast, token-free models \cite{xue2021byt5} operating on raw bytes could process any type of strings, making multilingual processing easy and reduces the complicated text processing pipeline. The trade-off is that converting raw texts to bytes generally result in longer input sequences, causing a slowdown in training and inference, especially for transformers that have a time complexity of $O(\log n^2)$. Yet byte-level encoding has been shown to benefit the multilingual G2P for at least a few languages \cite{yu2020multilingual}.

\textbf{Model architecture} Our G2P model is based on T5 architecture, a transformer encoder-decoder model that has been shown to be particularly powerful in a wide range of sequence-to-sequence tasks \cite{raffel2020exploring}. The current multilingual G2P model is based on the architecture of ByT5 \cite{xue2021byt5}, a variant of T5 operating directly on raw bytes. Transformers, including T5, have been shown to be particularly suitable for G2P in many languages \cite{yu2020multilingual,vesik-etal-2020-one,rezackova21_interspeech}. In addition to ByT5, we also tested the token-based mT5, the multilingual variant of T5 that has been pretrained on over 100 languages \cite{xue2021mt5}. The models were optimized for the standard crossentropy loss.

\textbf{Language prefixes} Language prefixes were appended before each word to specify the language, so that the model learns to perform G2P in individual languages. For example, if the word is the English word \texttt{yeet}, the actual input is \texttt{<eng-us>:yeet}.  In order to make the G2P model generalizable to unseen languages, we randomly sampled 15\% of tokens and appended \texttt{<unk>:} before them. As a result, the G2P model will learn a general correspondence between graphemes and phonemes, enabling it to perform zero shot prediction on unseen languages.

\begin{table*}[]
\centering
\caption{Evaluation results of multilingual G2P models. The generation speed was measured on an A40 GPU with a batch size of 512 without beam search and without batch decoding. The actual speech might fluctuate depending on various external factors but the relative ranking of generation speed should remain the same.}
\begin{tabular}{lrrrl}
\hline
Model  & PER (\%)  & WER (\%)  & Parameters & Speed \\ \hline
\multicolumn{3}{l}{{\it Pretrained weights}} &       \\ \hline
mT5-small       &  11.9    &   37.1  & 300M &  1500 words/second     \\
ByT5-small     &   \textbf{8.8}     &  \textbf{25.9} & 300M   &  917 words/second     \\\hline
\multicolumn{3}{l}{{\it Randomly initialized weights}} &       \\ \hline
mT5-tiny - 8 layers     &  28.9     &  70.8  & 67M  &  2250 words/second     \\
mT5-tiny - 12 layers       &  17.1    &   48.5 & 139M   &  2063 words/second     \\\hline
ByT5-tiny - 8 layers     &  9.8     &  28.8  & 7.3M  &  2475 words/second     \\
ByT5-tiny - 12 layers       &  9.7    &   28.3 & 15.9M   &  1833 words/second     \\
ByT5-tiny - 16 layers     &   \textbf{9.5}     &  \textbf{27.7}   & 20.6M & 1706 words/second     \\
\hline
\end{tabular}
\label{tab:main}
\end{table*}

\begin{figure*}[tbh]
    \centering

    \includegraphics[width=\textwidth, trim={0 0 0 0.7cm}, clip]{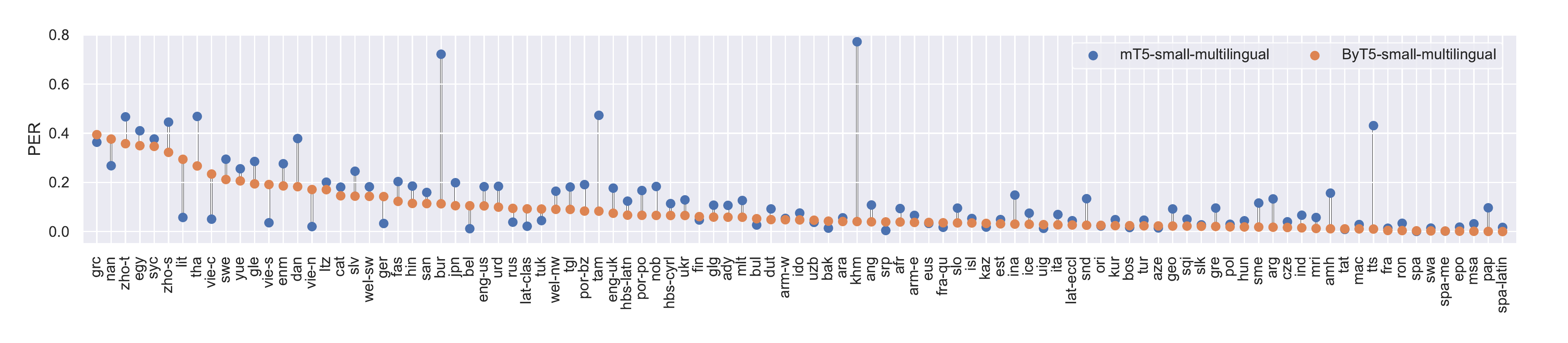}\vspace{-0.3cm}
    \includegraphics[width=\textwidth, trim={0 0.8cm 0 0.5cm}, clip]{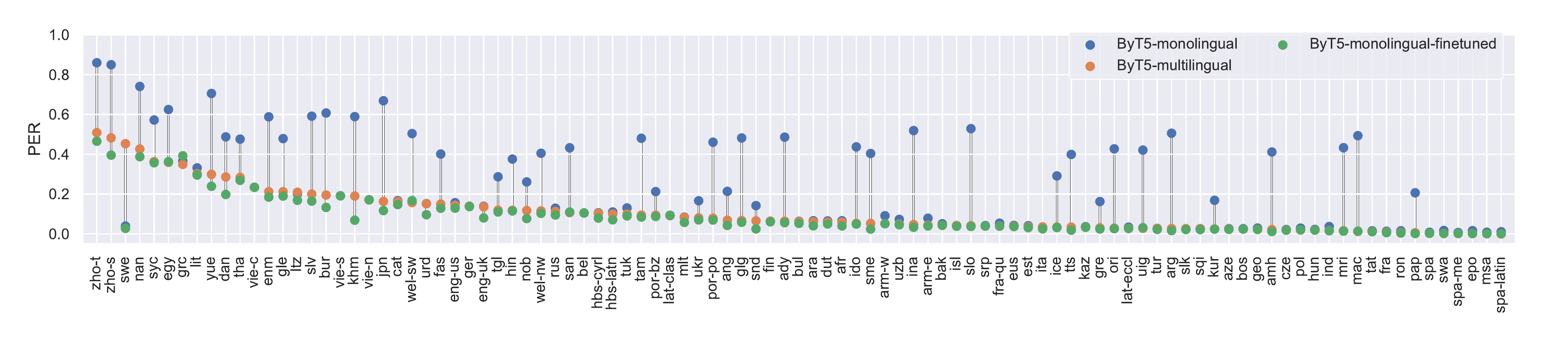}\vspace{-0.25cm}
    \caption{[Top]: Pairwise comparison of PER suggests that \texttt{ByT5-small} (orange dots) outperformed \texttt{mT5-small} (blue dots) in most languages. [Bottom]: The multilingual ByT5 (orange dots) has lower PER than monolingual ByT5 models (blue dots) in most languages. Yet monolingual ByT5 initiated with the pretrained multilingual ByT5 weights can further lower PER on a few languages (green dots). All of them are of the same \texttt{ByT5-tiny-8 layers} architecture).}
    \label{fig:compare}
\end{figure*}

\section{Experiments}
We trained a number of multilingual and monolingual models to perform G2P. All pretrained models, including their hyperparameters and training states, are available on the project website\footnote{\url{https://github.com/lingjzhu/CharsiuG2P\#pretrained-models}}. For multilingual models, we fine-tuned the pretrained \texttt{ByT5-small} and \texttt{mT5-small} on all language data. As baselines, we also trained lightweight version of the ByT5 model with 8, 12 and 16 layers. All models were trained on an A40 GPU with 48GB of memory. We set the effective batch size to 512 and the learning rate to $0.0003$ using the AdamW optimizer \cite{loshchilov2018decoupled}. All models were trained for 10 epochs and the validation PER was used to select final models. Generally, training multilingual models took about 0.8 to 2.5 days depending on the model size.

For comparison, we also trained two set of monolingual ByT5 models for each language on the same hardware. One set of models was trained with randomly initialized weights and the other set was fine-tuned from a pretrained multilingual model. All of these monolingual models were based on the same underlying architecture (\texttt{ByT5-tiny-8 layers}). All hyperparameters were the same as the multilingual models except that the batch size was set to 32. While this set of hyperparameters might not be optimal for all languages, we tried to make the comparison fair as both multilingual and monolingual models have seen the same amount of data for ten epochs. The training time for monolingual models ranged from less than 20 minutes for low-resource languages and more than 3 hours for high resource languages.

\begin{table*}[tbh]
\centering
\caption{Evaluation results on low resource languages. Results are presented in the format of PER/WER(\%).}
\begin{tabular}{lccccc}
\hline
\multicolumn{1}{c}{Model} & \multicolumn{5}{c}{Language}                                                                                                   \\ \cline{1-1} \cline{2-6}
      & Tibetan & Albanian & Hausa & Hebrew & Lower Sorbian \\ \hline
\multicolumn{1}{l}{{\it Zero-shot G2P}}   
\\\hline
 mT5-small     &   155.6 / 100     & 53.4 / 96        &    77.3 / 100         & 110.1 / 100            &   \textbf{38.8} / 89.5           \\ 
 ByT5-small     &   \textbf{109.3} / 100     & 56.1 / 99        &    76.9 / 100         & \textbf{101.8} / 100            &   42.7 / 89           \\ 
 ByT5-tiny - 8 layers     &   113.4 / 100     & \textbf{52.8} / 99        &    \textbf{75.9} / 100         & 137.8 / 100            &   45.1 / 93           \\ \hline
\multicolumn{1}{l}{{\it Finetuning}}                                   \\ \hline
 ByT5-small (text pretraining)     &  95.2 / 100          &   12.3 / 45   &          \textbf{25.4} / 78.5          &  36.6 / 89        &   7.6 / 35     \\
 ByT5-small (G2P pretraining)     &   \textbf{51.7} / \textbf{79}          &   \textbf{11.7} / \textbf{43.5}   & 25.9 / \textbf{76.5}          &  \textbf{36.1} / \textbf{84.5}            &     7.8 / 34.5               \\ 
 ByT5-tiny - 8 layers (G2P pretraining)     &   56 / 76.5     &  12.2 / 45    & 25.8 / 80      &  37.3 / 86.5     &     \textbf{6.4} / \textbf{30.5}               \\ 
 ByT5-tiny - 8 layers (random initialization)  & 73.2 / 94  & 51.1 / 90    &    62.6 / 97     &  58.5 / 94.5 &    31.2 / 74.5    \\ \hline
\end{tabular}
\label{tab:zero}
\end{table*}

\section{Evaluations}
During evaluation, we used beam search to generate the predicted pronunciations with a beam size of 5. Predicted pronunciations were measured by phone error rate (PER) and word error rate (WER) \cite{yu2020multilingual}. As seen in Table~\ref{tab:main},  byte-based ByT5 models outperformed the token-based mT5 models by a significant margin. Even for tiny ByT5 models that were randomly initialized, they still outperformed the mT5-small model that had more than 10 times of parameters and had been pretrained on massive texts. The fine-tuned \texttt{mT5-small} did not perform well on multilingual G2P, despite its large number of parameters. Because of its wordpiece tokenization, most of its parameters are word embeddings for 250k subword tokens, leaving fewer parameters for subsequent layers than ByT5. While \texttt{mT5-small} has achieved state-of-the-art in many multilingual NLP tasks \cite{xue2021mt5}, it might not be optimal for multilingual G2P. 
However, as byte representations are generally longer than token representations, ByT5 models tend to be slower than their mT5 counterparts in generation. 

A closer look at model performance for individual languages reveals that G2P models predict better in languages with Latin alphabet (see top panel of Fig~\ref{fig:compare}), as these languages tend to be more phonetically transparent. Languages with a phonetically regular spelling system like Spanish (\texttt{spa}), Esperanto (\texttt{epo}) and many European languages are easily handled by G2P models. Languages with non-alphabetical writing systems are generally harder to predict, such as the logogram-based Chinese language varieties (\texttt{zho,yue,nan}), Japanese (\texttt{jpn}), Vietnamese (\texttt{vie}), and most Asian languages. 

Multilingual training also improves the overall performance of G2P, as monolingual models generally have higher PER and WER in comparison with the multilingual model matched in architecture and parameters (see bottom panel of Fig~\ref{fig:compare}). While this strongly suggests the benefits of multilingual pretraining, it is also noted that ByT5 models are harder to train and converge, so careful hyperparameter tuning could still improve the results of monolingual models. Finetuning the multilingual model on individual languages further reduces PER, and this improvement is more evident in non-Latin based languages.

\begin{figure}[tbh]
    \centering
    \includegraphics[width=0.48\linewidth]{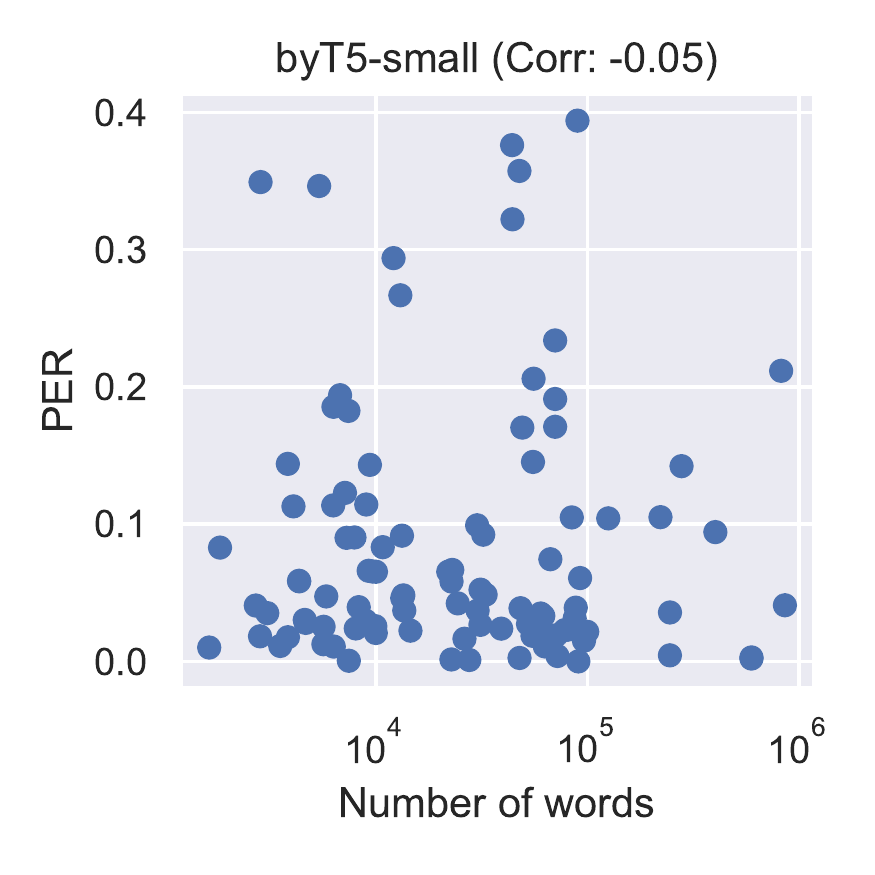}
    \includegraphics[width=0.48\linewidth]{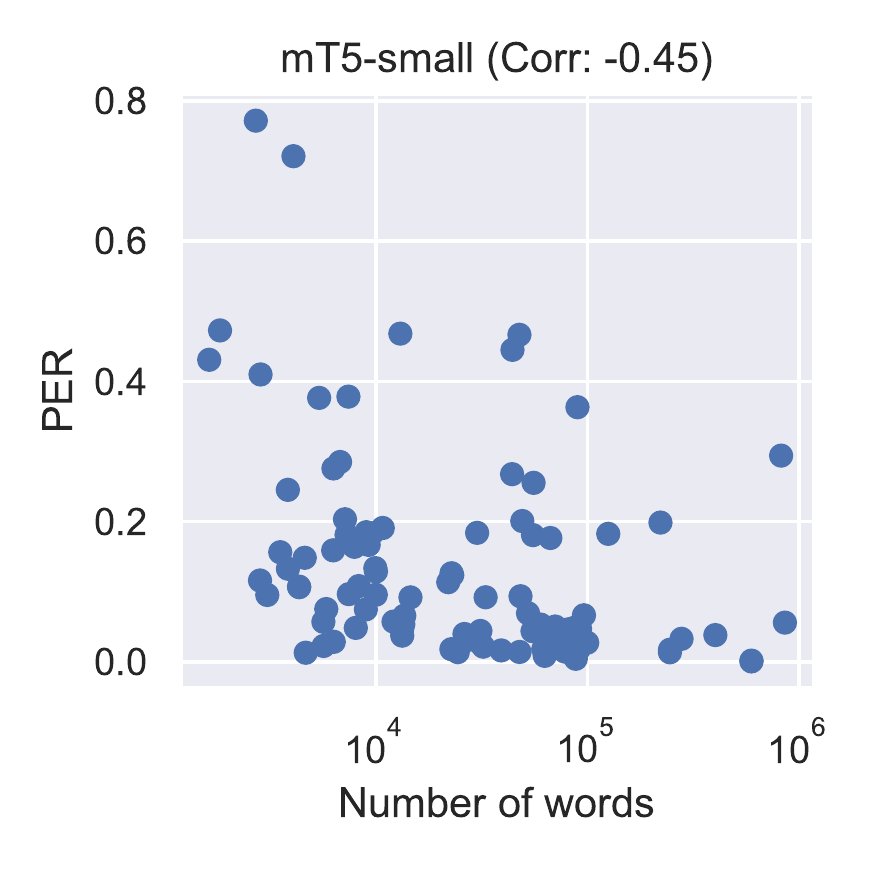}
    \caption{For \texttt{byT5}, the PER is not directly correlated to the dictionary size of training languages (Spearman's $\rho$: -0.05), whereas for \texttt{mT5}, the PER has a moderate negative correlation with the size of training languages (Spearman's $\rho$: -0.45).}
    \label{fig:corr}
\end{figure}

We analyzed the relationship between the data size and the final performance (Fig~\ref{fig:corr}). For mT5, data size has a moderate negative correlation with the final performance (Spearman's $\rho$: -0.45), as more data tend to lower the PER in that language. However, there is almost no correlation between data size and PER for multilingual ByT5 models (Spearman's $\rho$: -0.05). For monolingual models, PER is negatively correlated to the size of dictionary (Spearman's $\rho$: -0.64) if these models are trained from randomly initialized weights. Yet if these monolingual models are fine-tuned, the correlation is almost non-existent (Spearman's $\rho$: -0.10). 

\section{Low resource G2P}\label{low}
While ByT5 works well for language seen in training, we also tested whether the multilingual G2P models can generalize to unseen low resource languages and whether low resource languages can benefit from finetuning on pretrained weights. 

The zero-shot experiments were conducted on five typologically diverse languages, Hebrew (\texttt{heb}), Tibetan (\texttt{tib}), Hausa (\texttt{hau}), Albanian (\texttt{alb}) and Lower Sorbian (\texttt{dsb}). Among these languages, Hebrew and Tibetan have their own writing systems, whereas the rest of languages use the Latin alphabet. For each language, we partitioned 50 words for model selection and 200 words for testing, leaving the rest of the data for training (1500-2000 words). Neural models for these languages were trained in the same method as other monolingual models. 

Results in Table~\ref{tab:zero} show that both ByT5 and mT5 models can perform zero-shot prediction on unseen languages in Latin alphabet, as the Latin alphabet is often phonetically transparent and is used by the majority of languages in the training data. The multilingual G2P models can generalize to unseen languages based on the learned word-sound correspondence from other languages with a similar writing system. Yet the genalizability is highly limited, as the models can only predict certain phones but cannot correctly predict word-level pronunciations. For Hebrew and Tibetan, unsurprisingly, none of the models can predict pronunciations with unseen writing systems. 

Given that zero-shot prediction does not always generalize, we tested whether fine-tuning pretrained multilingual models can further reduce PER. Specifically, we compared whether weights from a pretrained text model (the original \texttt{ByT5-small}) and weights from pretrained G2P model can benefit low resource G2P. Even with only about 1800 training words in each language, finetuning can successfully reduce both PER and WER by a significant margin.
If the model was initialized with \texttt{ByT5-small} pretrained weights, not only did the model generally converged to higher PER and WER but also took more iterations to train. However, for low resource languages, the training dynamics of G2P models was very unstable, with PER rate varying greatly at successive checkpoints and susceptible to overfitting.

\section{Conclusions}
Amassing around 100 publicly available grapheme-to-phoneme datasets, we trained large-scale multilingual G2P models based on ByT5.  We found that byte-level transformers significantly outperformed the token-based mT5 and monolingual models. The pretrained multilingual G2P model can benefit low resource G2P as it can perform zero-shot prediction on certain unseen languages. We also demonstrate that finetuning monolingual data on pretrained multilingual model generally helps the model converge to lower PER than randomly initialized models.

However, it should be noted that statistical G2P models are not the optimal model for \textit{every} language. For logogram-based Chinese, querying from a pronunciation dictionary is more effective than using a neural G2P model at the word-level, though phrase-level G2P is still needed for Chinese to disambiguate homographs \cite{Park2020}. For phonetically regular languages like Spanish and Italian, rule-based methods could work very well. Yet for the majority of writing systems that are only partially phonetically regular, neural G2P models can effectively generate phonemic transcriptions from texts. In the future, we will improve our multilingual G2P toolkit with linguistic knowledge, so as to include more languages and improve its performance.

\newpage
\bibliographystyle{IEEEtran}

\bibliography{mybib}

\end{document}